\documentclass[11pt,letterpaper]{article}

\usepackage[T1]{fontenc}
\usepackage{lmodern}
\usepackage[margin=1in]{geometry}
\usepackage{amsmath,amssymb}
\usepackage{graphicx}
\usepackage{booktabs}
\usepackage{microtype}
\usepackage{xcolor}
\usepackage{tikz}
\usepackage{float}
\usepackage{needspace}
\usepackage[subpreambles=true]{standalone}
\usepackage[font=small,labelfont=bf]{caption}
\usepackage[numbers,sort&compress]{natbib}
\usepackage{xurl}
\usepackage[hidelinks,pdfusetitle]{hyperref}
\hypersetup{pdfauthor={Yukun Dong, Linjun Zhang, Tian Zheng}}

\providecommand{\Description}[2][]{}

\newcommand{\STAIX}{STAI-X}
\newcommand{\STAIXYear}{2026}
\newcommand{\STAIXConf}{\STAIX~\STAIXYear}
\newcommand{\checkfield}[1]{\textit{#1}}

\title{Author-in-the-loop Agentic Reproducibility Auditing for a Machine Learning Conference}

\author{Yukun Dong$^{1}$\footnote{authors are listed in alphabetical order}, Linjun Zhang$^{1}$, Tian Zheng$^{2}$\\[1.2ex]
 {\normalfont\small $^{1}$Rutgers University\quad $^{2}$Columbia University}}
\date{}

\begin{document}

\maketitle

\begin{abstract}
Reproducibility checks during peer review require a shared understanding of what authors' code is expected to reproduce. We present an author-in-the-loop collaborative audit that makes author declarations explicit inputs to an AI agent's investigation and assessment. A structured checklist records reproduction instructions and limitations; a two-stage workflow on GitHub connects agent checks to author clarification, revision, and agent reassessment. The agent separately assesses whether results have been reproduced and whether authors need to provide clarification or make repairs, while human reviewers approve requests and decide whether concerns are resolved. We deployed the system across 36 conference papers and 133 experimental tables and figures, raising 20 issues after human review. Further checks supported closing all the issues, although full reproduction remained unverified in some cases. We further compared repeated audits with complete and blank checklists to examine how author declarations shaped the agent's investigation and assessment. Our work contributes a collaborative workflow and agent design that use author knowledge to guide reproducibility auditing. The conference deployment provides empirical evidence that this approach can help identify and address reproducibility problems through collaboration among authors, agents, and reviewers.
\end{abstract}


\begin{figure}[htbp]
    \centering
    \includegraphics[width=\textwidth]{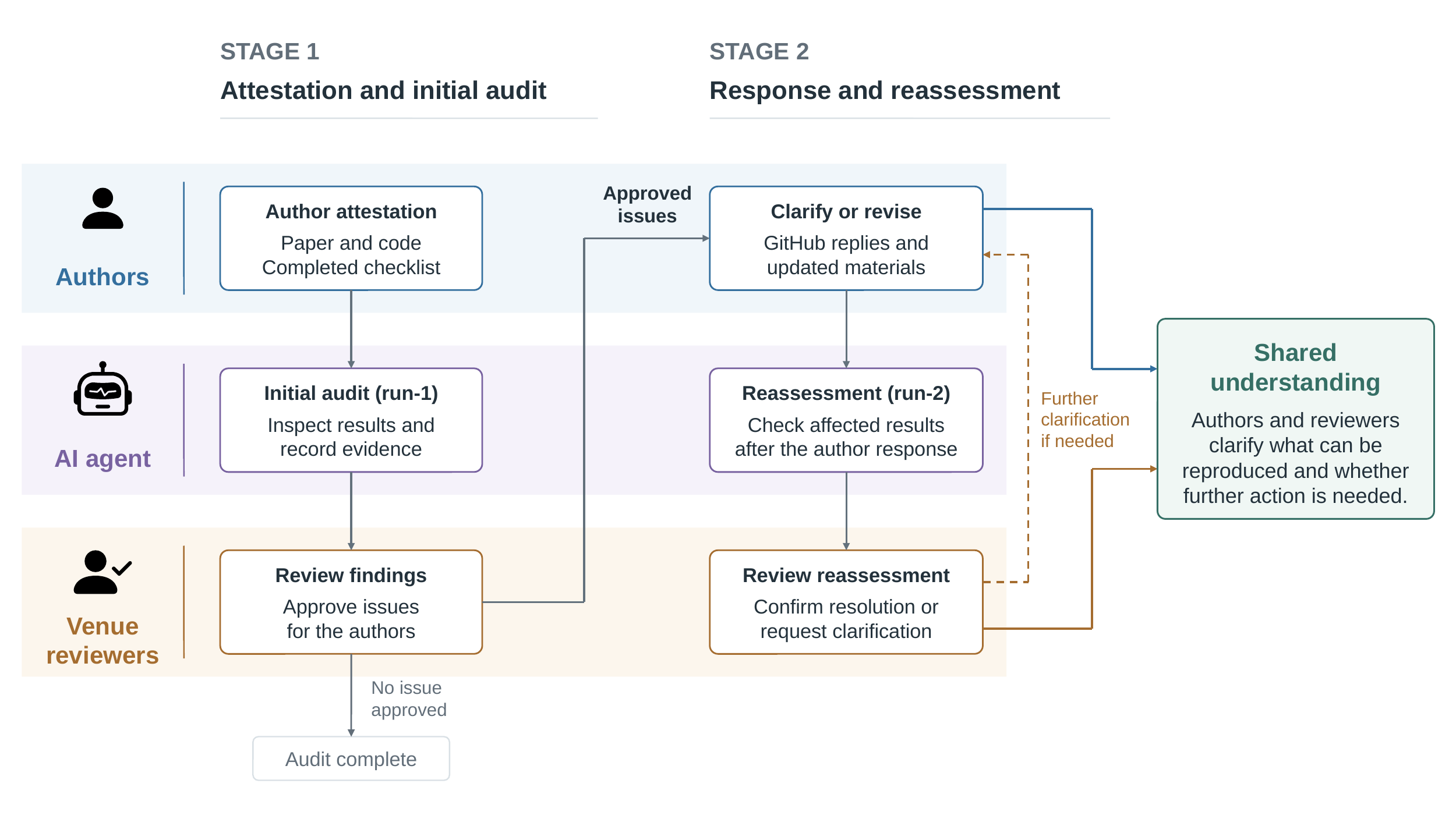}
    \caption{The two-stage collaborative audit workflow. In the first stage, the author's checklist guides the
    initial agent code check. Papers with human-approved issues enter the second stage for response and
    reassessment, with human reviewers deciding issue resolution. The workflow
    aims to help authors and reviewers clarify what can be reproduced and what
    requires clarification or repair.}
    \Description{Three horizontal lanes represent authors, the AI agent, and
    venue reviewers. In Stage 1, authors submit their paper, code, and completed
    checklist. The agent performs an initial audit and human reviewers assess
    the findings. If no issue is approved, the audit ends. Approved issues
    lead to Stage 2, where authors clarify or revise their materials, the agent
    reassesses affected results, and human reviewers decide resolution. A loop
    returns remaining questions to the authors. Paths from the author and
    reviewer lanes converge on the collaboration goal of clarifying what can
    be reproduced and whether further author action is needed.}
    \label{fig:collaborative-workflow}
\end{figure}

\section{Introduction}
\label{sec:introduction}

Reproducibility is central to rigor and advancement in computational scientific research:
 obtaining consistent results from the original data, code, and analysis procedures 
 enables researchers to evaluate published findings and build on prior work. To support this goal, journals have introduced dedicated reproducibility reviews, often as a final check after acceptance, 
 that examine the computational materials accompanying a manuscript~\cite{jasa2024partnering,nuest2021codecheck},
  alongside related reproducibility initiatives at conferences~\cite{pineau2021improving}. 
  These efforts bring code, data, and computational workflows into the publication process. 

However, a paper and its code may not give a reproducibility reviewer sufficient information to assess the reported results. Reproducibility checks are complicated, as authors often know relevant details that are not explicit in
the paper. The submitted code may use a different implementation of a baseline from the one used in the paper. A script may recreate a table by reading numerical results already saved by the authors, without rerunning the experiments that produced those numbers. These details affect how reviewers interpret their
checks and decide whether a problem requires correction. Without this
context, avoidable misunderstandings can arise between authors and
reviewers, and important code problems may remain unresolved. This
information asymmetry between authors and reviewers can make the review process inefficient, which motivates our first research question (RQ): 

\noindent\textbf{RQ1.} How can a venue structure collaboration between authors
and reviewers to reduce information asymmetry in reproducibility auditing?

To address RQ1, we design an author-in-the-loop collaboration process with
two stages, combining \emph{structured author declarations} with a
\emph{transparent communication workflow} to help bridge the information gap
between authors and reviewers. The first stage begins with authors completing a
structured checklist that records relevant details for each experimental
figure or table, such as uncertainty in the results and estimated
code execution time. This gives the venue's reviewers explicit statements from
the authors about how each result is obtained and what may limit its
reproduction. Authors who receive issues identified by reviewers then enter
the second stage, providing clarifications or revised materials for
reassessment. A shared workflow on GitHub makes the audit process visible
to participants and supports timely communication throughout both stages.

Our emphasis on collaboration between authors and reviewers builds on prior work in reproducibility review~\cite{jasa2024partnering,nuest2021codecheck}, but conducting these reviews entirely through human effort is resource intensive.
Recent advancements in AI agents~\cite{openaiCodexCLI,anthropicClaudeCode} offer new capabilities to mitigate this tension by checking the submitted code and running experiments to compare their outputs with the results reported in the paper. However, recent work shows that reproducibility agents encounter diverse
outcomes that require careful interpretation~\cite{baek2026automated,wu2026artifact}.
Full code execution may exceed the resources available to a venue, while a failed
run may instead reveal an error in the submitted code. Further care is
needed when an agent uses these findings to request revisions from authors:
each request needs to identify something the authors can reasonably address.
For example, data may be unavailable because of privacy
restrictions~\cite{krafczyk2021learning}, or authors may explicitly provide
only a simplified demonstration because the full experiment takes too long
to run. These limitations may prevent full reproduction, but they do not by
themselves justify asking authors to revise their materials. Author declarations
therefore provide important context for interpreting execution outcomes and
deciding whether clarification or repair is warranted. This motivates our
second research question on incorporating author declarations into the
agent's assessment process:

\noindent\textbf{RQ2.} How can an AI agent use author declarations and responses
to interpret complex reproduction evidence and make auditable judgments about
the need for clarification or repair?

Accordingly, we design the agent to treat author declarations and responses as explicit inputs to its investigation and assessment. They guide the scope of checking and inform the interpretation of its outcomes. For each experimental figure or table in the paper, we define a multidimensional evidence state that distinguishes execution status, diagnostic findings, and result agreement with the paper to deal with heterogeneous code execution situations. A separate verifier agent checks this evidence and uses the authors' checklist declarations to judge whether clarification or repair is warranted under the venue's audit conditions. We preserve execution records and assessment revisions so that authors and reviewers can check the evidence and reasons behind each judgment.

The practical value of author-in-the-loop auditing also depends on the problems it identifies and whether they can be addressed. To examine these outcomes, we deployed our collaborative protocol and agent at \STAIXConf, covering 36 papers before publication. We traced each concern through author responses and further checks to examine whether and how it was addressed. This deployment provides the basis for our third, empirical research question:

\noindent\textbf{RQ3.} What reproducibility problems does the pipeline identify
in a conference deployment, and to what extent are they resolved after author
responses and further checks?

At \STAIXConf, our deployment examined 133 experimental results across 36 papers
and raised 20 issues across ten papers after human review. These included
missing implementations, execution errors, and discrepancies between the
submitted code and the paper. Authors responded with code revisions,
additional materials, and clarification of what their releases were intended
to reproduce. Subsequent checks and human review supported closing all 20
issues, although full reproduction remained unverified in some cases.
We first examined how execution outcomes related to judgments about the
need for author action. In the initial audit, 68 of 85 unsuccessful checks
were judged not to warrant a request, while three of 48 successful checks
still warranted clarification or repair. These findings provide a quantitative
account of how code execution outcomes relate to judgments about author action
in a conference audit. We then conducted comparative
experiments on the ten papers that received issues, showing that author
declarations influenced these judgments. The deployment and comparative experiments illustrate how author participation can inform agent assessments and help resolve concerns raised during an audit.

In summary, this paper makes three contributions:

\begin{itemize}
\item \textbf{An author-in-the-loop reproducibility audit protocol.} We combine a structured
checklist for individual experimental results with a two-stage collaborative
workflow designed to reduce information asymmetry between authors and
reviewers. The protocol connects author declarations to audit feedback and
makes subsequent responses and review decisions traceable.

\item \textbf{An auditable agent that incorporates author input.} We make
author declarations and responses explicit inputs to the agent's investigation
and assessment rules. The agent interprets reproduction evidence in light of
this information while separately judging whether the findings warrant
clarification or repair, retaining the evidence and reasons for reviewers to
examine.

\item \textbf{Empirical insights from a conference deployment.}
We report the types and frequency of reproducibility problems identified
during the conference deployment. We also quantify how code execution
outcomes relate to judgments about author action and trace how author
responses and human review contribute to issue resolution. Comparative
experiments further examine how author declarations shape the agent's
assessments.

\end{itemize}

The rest of paper is organized as follows. After reviewing related work in Section~\ref{sec:related-work}, we present the author checklist and two-stage collaborative workflow in Section~\ref{sec:collaborative-audit-protocol} to address RQ1. Section~\ref{sec:agent-design} addresses
RQ2 through the agent design, explaining how author input guides its checks
and how it interprets reproduction findings before recommending whether
author action is warranted. Section~\ref{sec:findings} examines all
three research questions through deployment records and comparative experiments.
We conclude in Section~\ref{sec:discussion}
with the study's limitations and implications for reproducibility checking
in peer review.

\section{Background and Related Work}
\label{sec:related-work}

We review related work from two perspectives: the design of reproducibility
audits and empirical findings from their evaluation and deployment. The first
concerns how authors, reviewers, and AI agents participate in the checking
process; the second considers what prior observations and experiments reveal
about how these audits work in practice.

\subsection{Design Approaches: Incorporating Author Input into Reproducibility Auditing}

For journals and conferences, designing reproducibility audits with AI
assistance involves two related questions: how to organize collaboration
between authors and reviewers, and how to design the agents that support
their work.

Author declarations help reviewers establish what the submitted code is
intended to reproduce. The NeurIPS checklist asks authors to describe
experimental procedures and computing resources~\cite{pineau2021improving}.
The Journal of the American Statistical Association (JASA) further asks which
figures and tables the code reproduces and what restrictions affect data
access or execution~\cite{jasaACCForm}. EuroSys artifact appendices connect
paper claims to experiments and specify expected outputs and resource
estimates~\cite{eurosys2025artifactappendix}.
Some NeurIPS reviewers reported that checklist answers helped them assess
submissions~\cite{pineau2021improving}.
Author declarations guide the initial checks, while continued communication
allows authors and reviewers to clarify findings and address problems
encountered during the audit. CODECHECK and
JASA incorporate communication with authors to resolve uncertainties and
address problems~\cite{nuest2021codecheck,jasa2024partnering}. The MLSys 2026
deployment extends this practice to agent checks followed by author
clarifications or repairs and further checking~\cite{yang2026aeagent}.

At the level of agent design, prior studies highlight the difficulty of
interpreting complex code execution outcomes~\cite{baek2026automated,wu2026artifact}.
A check may reproduce only some entries in a table, or recalculate its
values from data supplied by the authors without rerunning the experiment
that generated those data~\cite{baek2026automated}.
Conversely, a failed run may reflect problems in the checking environment
rather than errors in the authors' code~\cite{wu2026artifact}.
Several systems have introduced explicit checks to help interpret these outcomes.
Artisan checks whether an AI agent calculates the results or simply copies
values already stored in the repository~\cite{baek2026automated}.
ArtifactCopilot checks whether the outputs obtained by running the code
support specific methods or results described in the paper. When only
part of the experimental workflow is run, it requires an explicit
statement of what was checked~\cite{wu2026artifact}.
To make agents' judgments traceable,
PaperRepro and VERITAS preserve execution logs and records of code
changes~\cite{zhang2026paperrepro,liu2026veritas}.

Another design consideration is how agents use author declarations during reproducibility checks. The NeurIPS Checklist Assistant evaluates
authors' checklist answers and explanations against the manuscript,
but does not run the code~\cite{goldberg2024checklist}.
The Artisan study did not count differences between code outputs and paper
results as newly discovered problems if the authors of the audited papers
had already disclosed them~\cite{baek2026automated}.
The study does not describe how the agent uses these author explanations
to decide whether to request clarification or code changes during the reproducibility audit.

Compared to existing approaches, our design makes author declarations explicit inputs to the agent's code check
and execution, and to decisions about requesting author for further actions.
For each experimental result, the agent considers whether an observed
problem is explained by a declared limitation or conflicts with what the
authors promised to provide, giving a basis for requesting clarification
or repair under the venue's requirements.
We collect these declarations in a checklist for experimental figures and
tables and organize checking and author responses through a two-stage
collaborative workflow on GitHub. Shared GitHub records link requests
and review decisions to the supporting evidence, with human reviewers
deciding which issues to raise and close.

\subsection{Empirical Evidence: Audit Outcomes and the Influence of Author Input}

Research on AI-assisted reproducibility checking has yielded a wide range of empirical observations. In the Artisan
study, authors confirmed discrepancies between papers and computational
materials and repaired identified code errors~\cite{baek2026automated}.
The ICML 2026 Open Reproductions initiative similarly reported author
confirmations and paper corrections. In one case, organizers withdrew a
claim that a paper's result was incorrect after finding that the reproduction
compared runtimes expressed in different units~\cite{abid2026icmlreproductions}.
At MLSys 2026, authors
repaired or clarified problems identified by the agent, which then checked
their responses. For some papers, the agent reviewed training logs supplied
by the authors without rerunning the full training
process~\cite{yang2026aeagent}. These studies show that reproducibility
audits can prompt authors to correct their papers or code. However,
mistakes made during an audit can also lead auditors to wrongly conclude
that a paper's results are incorrect.

Several studies have also reported experimental findings on AI-assisted
reproducibility checking.
ArtifactCopilot shows that detailed review guidance can improve agreement
with human reference judgments, although conclusions still vary across
repeated audits~\cite{wu2026artifact}. Additional information can also bias
judgments. In SocSci-Repro-Bench, providing the paper PDF modestly improved
overall accuracy, but agents sometimes returned numbers from the paper when
missing materials prevented reproduction~\cite{alizadeh2026socsci}.
Checklist explanations can also influence evaluations based on manuscript
text: in an adversarial test of the NeurIPS Checklist Assistant, rewriting explanations for
checklist answers made evaluations more favorable while the manuscripts
remained unchanged~\cite{goldberg2024checklist}.

Prior studies describe problems identified by agents and changes made by
authors, but provide limited quantitative evidence on how agents' code execution outcomes
relate to requests for authors' actions. We examine this relationship
in our deployment at \STAIXConf{} and use comparative experiments to
investigate how author declarations influence these decisions. We also
examine author responses and the agent's reassessments to understand how the issues were addressed.
The experiments compare agent runs using completed checklists with runs using blank templates, showing that checklist answers can influence which code agents inspect or run and whether they ask authors for clarification or corrections.

\section{Designing a Collaborative Author-in-the-loop Reproducibility Check Workflow}
\label{sec:collaborative-audit-protocol}

In this section, we present a collaborative author-in-the-loop workflow designed to reduce information asymmetry between authors and reviewers during reproducibility checks, which is our first research question.
This pipeline coordinates authors, the AI agent, and human reviewers through two stages
guided by the venue's reproducibility requirements. In the first stage, author declarations
provide the starting point for agents' first run (run-1, initial audit).
Papers with issues approved by reviewers enters the second stage, where author responses guide agents' second run (run-2, reassessment), after
which human reviewers determine whether the concerns have been adequately
addressed. We deployed this workflow at \STAIXConf, where accepted papers
underwent reproducibility checking before publication and unresolved issues
prevented their inclusion in the proceedings.

\begin{figure}[tbp]
    \centering
    \includegraphics[width=\linewidth]{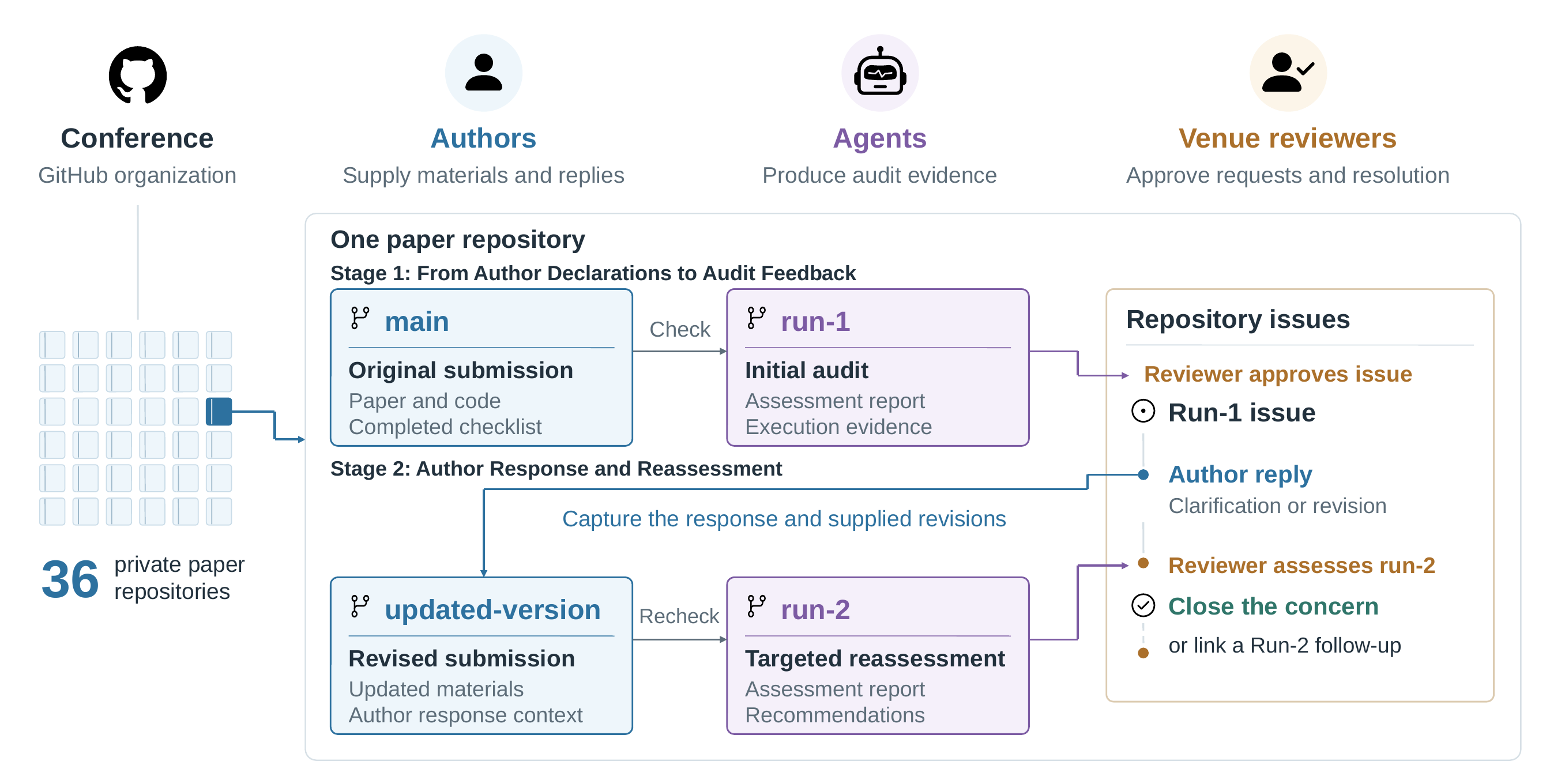}
    \caption{GitHub organization and audit records supporting the two stages
    of collaborative reproducibility checking.}
    \Description{A generic conference heading and GitHub logo appear above a
    collection of 36 paper repositories. One repository is expanded into two rows of
    branches and a shared issues area. Stage 1 covers author declarations
    through audit feedback: original materials in main feed the initial
    audit in run-1, and a reviewer approves an initial issue. Stage 2 begins
    with the author reply and covers response and reassessment. For papers
    with approved issues, updated-version stores
    the supplied revisions and response context, which feed the targeted
    audit in run-2. Reviewers assess the recommendation and close the concern
    or link a run-2 follow up issue. Person and robot icons identify authors,
    agents, and venue reviewers above their respective contributions.}
    \label{fig:github-audit-management}
\end{figure}

GitHub provided a shared record across both stages as shown in Figure~\ref{fig:github-audit-management}. A conference organization housed
36 private repositories, one per paper, with separate branches
(\texttt{main}, \texttt{run-1}, \texttt{updated-version}, and \texttt{run-2})
for submitted and revised materials and their corresponding agent checking outputs. GitHub issue histories preserved author responses and the
reasons for retaining or closing concerns, allowing authors and reviewers to
trace decisions back to the evidence. 

\subsection{Stage 1: From Author Declarations to Audit Feedback}

In Stage~1, authors complete a checklist that guides the agent's initial auditing of
their submitted code and data. The agent then conducts checking and proposes GitHub issues requesting
clarification or repair, subject to human reviewer approval.

\paragraph{Designing the Author Checklist.}
\label{sec:author-checklist}

We design the checklist to make explicit which results the submitted code
and data are intended to reproduce and under what conditions. Authors'
completed answers in the checklist, referred to here as author declarations or self-attestations, guide both agent's
computational checks and requests for author action.
The checklist has two sections. The opening questions section asks authors general reproducibility questions about the paper, the execution table section asks for one row per experimental figure or table
in the main text. The execution table's fields address the following needs:

\begin{itemize}
    \item \textbf{Connecting paper's results to an execution path.}
    Each row in the execution table identifies a figure or table result in the submitted paper through its
    \checkfield{Paper locator}.
    The \checkfield{Entrypoint} gives the script or notebook path, and the
    \checkfield{Canonical command} specifies the command to run.
    These fields give the agent a starting point for locating the computation
    behind the reported result. 

    \item \textbf{Making expected agreement explicit.}
    The \checkfield{Result description and tolerance} and
    \checkfield{Expected output and runtime} fields specify the outputs
    expected from the declared command. Authors state tolerances or sources of randomness
    when exact agreement is not expected, giving the agent a reference for
    examining numerical differences in context.

    \item \textbf{Declaring what can be reproduced.}
    The \checkfield{Data availability} and \checkfield{Reproducibility} fields
    distinguish access to the required data from how much of a reported
    result the submitted materials can reproduce. Authors declare Yes,
    Partial, or No for each result. Partial indicates that only part of the
    result or a reduced version can be reproduced, such as a demonstration
    with synthetic inputs when the original data are restricted.
    Partial and No entries require explanation of omitted
    or blocked components; Yes and Partial require an expected output and
    runtime. These statements help delimit requests for clarification or
    repair. 

    \item \textbf{Exposing computational and service constraints.}
    The \checkfield{Expected output and runtime} and
    \checkfield{Notes on computation and stability} fields describe conditions
    such as extended training or live API dependencies. These declarations help the agent assess what can be checked within the audit budget set by the venue and explain what remains unverified. 
\end{itemize}

\paragraph{From Declarations to Reviewed Findings.}
\label{sec:synthetic-demonstration}

The initial audit (run-1) starts from the reproduction instructions and
constraints recorded in the completed checklist. The agent examines the
submitted code against the paper and, where execution is feasible, compares
generated outputs with the reported results. Its code checking report documents the checks
performed and identifies results that remain unverified. For problems that
may warrant author action, the agent drafts issues explaining why
clarification or fix
 is needed in light of the authors' checklist answers. The details
 of the agent design is discussed in Section~\ref{sec:agent-design}.

All issues proposed by the agent undergo human review.
Approved requests are posted as open GitHub issues for authors to review
and respond to. Each issue connects the affected result to supporting
evidence and a specific request. Only papers with approved issues enter
the required response and reassessment stage (run-2).

\begin{figure}[!htbp]
    \centering
    \includegraphics[width=\linewidth]{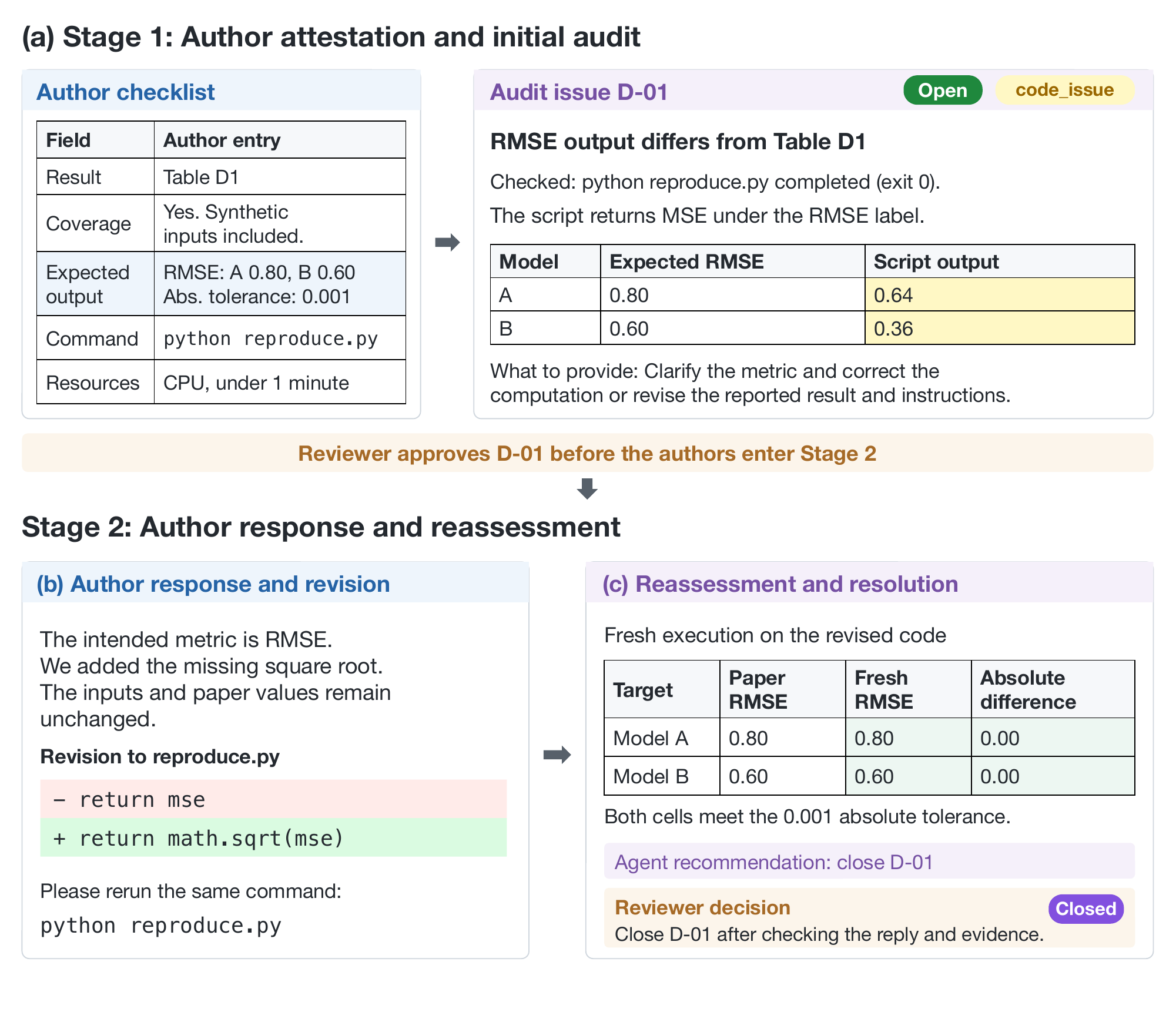}
    \caption{Synthetic demonstration of the two-stage collaborative audit.
    (a) Author declarations guide the initial check, and a reviewer approves
    the issue. (b) The author clarifies the metric and revises the code.
    (c) Fresh output supports a recommendation to close the issue, which a
    reviewer accepts.}
    \Description{A composite illustration presents Stage 1 across the top
    and Stage 2 in two panels below. In panel a, a synthetic author checklist
    declares Table D1 reproducible with python reproduce.py, an absolute
    tolerance of 0.001, and expected RMSE values of 0.80 and 0.60. The script
    instead returns 0.64 and 0.36 because it reports mean squared error as
    RMSE. Issue D-01 requests clarification or correction, and an illustrated
    reviewer approval leads to Stage 2. In panel b, the example author reply
    explains that the inputs and targets remain unchanged while the code
    adds the missing square root. An arrow leads to panel c, where fresh
    values of 0.80 and 0.60 match both targets. The agent recommends closing
    D-01, and an illustrated reviewer decision accepts closure. The demonstration
    covers only these two cells and does not establish whole paper
    reproducibility or represent an observed deployment interaction.}
    \label{fig:case-a-collaborative-audit}
\end{figure}

Figure~\ref{fig:case-a-collaborative-audit} illustrates both stages with a
constructed example.\footnote{We use a simplified synthetic example to preserve
anonymity, as details from an actual case could identify the submission or
its authors. The conference audit involved substantially more complex
cases.}
In panel (a), Table~D1 in the example paper reports root mean squared
error (RMSE) for two models. The author's checklist states that
\texttt{python reproduce.py} reproduces Table~D1 from included synthetic
inputs, with an absolute tolerance of $0.001$. The agent runs this script
successfully but finds the outputs mismatch the paper's results. It produces $0.64$ for
Model~A and $0.36$ for Model~B, whereas Table~D1 specifies $0.80$ and $0.60$.
The issue (labeled as D-01) connects this discrepancy to the checklist declaration and
requests clarification or correction of the computation and reported result.
The example then shows a reviewer approving the request before the author
responds and the revised code is checked in Stage~2.

\subsection{Stage 2: Author Response and Reassessment}
If there are open GitHub issues raised in Stage 1, the audit enters Stage 2.
In Stage 2, the agent checks whether authors' revisions address the original concerns, and human reviewers decide whether the issues can be closed. They can clarify what their submitted materials are intended to reproduce or revise the materials relevant to the issue. For the second audit (run-2), each response and any revised materials are recorded alongside the original issue and initial audit report. Authors and reviewers can then identify what has changed and which concerns the reassessment needs to address.

After authors' revision, the
agent checks the author's response against the updated materials and performs
further execution when needed. The run-2 agent returns a recommendation of whether to keep the issue open or mark as resolved with
supporting evidence, and human reviewers decide whether the issue has been
adequately addressed. In figure~\ref{fig:case-a-collaborative-audit}(b), the author's acknowledges that the script returned mean squared
error under an RMSE heading. It describes a revision that takes the square
root of the mean squared error and affirms that the intended targets remain
$0.80$ and $0.60$. This clarification
connects the requested repair to the metric defined in Table~D1 and specifies
the calculation to be checked again. The run-2 agent reran the corrected calculation and the resulting RMSE values, $0.80$ and $0.60$, match the two specified
cells within the declared tolerance. Figure~\ref{fig:case-a-collaborative-audit}(c)
pairs this computed evidence with an agent's recommendation to close issue D-01 and human reviewer agreed to mark it as resolved.

Across the two stages of collaboration, the checklist translates the venue's reproducibility requirements into explicit statements about what each submission is intended to reproduce and under what conditions. The GitHub workflow gives authors and reviewers a shared record for examining those statements and addressing concerns through author's revisions. This design helps authors and reviewers work toward agreement on what needs to be reproduced and what authors need to clarify or revise. Both can consult the GitHub records to see why reviewers raise or close each issue.

\section{Reproducibility Checking Agent Design}
\label{sec:agent-design}

The previous section describes the collaborative process design, but treating the code checking agents as black boxes. Here, we focus on the agent design in answer to our second research question, especially discuss how the agent uses author declarations to guide its checking and assessment. 

Prior reproducibility checking agents link execution records to assessments of reproduced
results~\cite{zhang2026paperrepro,baek2026automated,wu2026artifact}. Author
replies also guide further checks in the MLSys
deployment~\cite{yang2026aeagent}. Building on these approaches, our design makes author declarations explicit
inputs to the agent's rules for checking code and requesting clarification
or repair. Author's declarations can be potentially helpful: they can let an agent identify the code to check and plan feasible execution. They can also provide context for interpreting execution outcomes and deciding whether authors need to clarify or revise their materials. Once an issue is raised, author replies and revisions offer new information for reassessing whether the issue has been addressed. These uses of author
input motivate our first three design requirements; the fourth concerns the auditability of the checking process. We define the four design requirements (DRs) as follows:

\begin{itemize}
  \item \textbf{DR1. Code checks should use author declarations as guidance.}
  In a large repository, locating the code for a particular figure or table
  can require substantial investigation. Even after finding the relevant
  code, the agent needs to estimate runtime and assess whether execution is
  feasible within the audit budget. Author declarations could therefore
  guide both code identification and execution planning. Even the agent is
  given the author's declarations, it still needs to check whether the code implements the
  computation described in the paper correctly.

  \item \textbf{DR2. Agents need to consider author declarations when interpreting outcomes and justifying requests.}
  An incomplete reproduction by agent does not necessarily mean that authors
  need to revise their materials. The checklist may explain that private data
  limit the available experiments to a demonstration. However, if authors
  state that the required code is included but it is missing from the
  repository, the agent may ask them to provide it. The agent therefore needs
  to consider author declarations when interpreting execution outcomes and
  explain separately why a finding does or does not require clarification or
  repair.

  \item \textbf{DR3. Reassessment requires checking author responses.}
  Authors may respond to an issue by correcting their code or clarifying
  details that the agent misunderstood. These responses can change whether
  the original request is still justified. The agent should therefore check
  the authors' explanations and any revisions against the code and data,
  then explain why it recommends keeping the issue open or closing it.

  \item \textbf{DR4. Agent checks and judgments must remain auditable.}
  The agent runs must be traceable to the commands executed and the outputs
  obtained. Any changes during the audit must remain visible to
  authors and reviewers.

\end{itemize}

We implement these requirements as an agent skill, a reusable package
of task instructions and supporting scripts~\cite{zhang2025agentskills,openaiAgentSkills}. The system follows a classical orchestrator-worker pattern~\cite{anthropic2024building}.

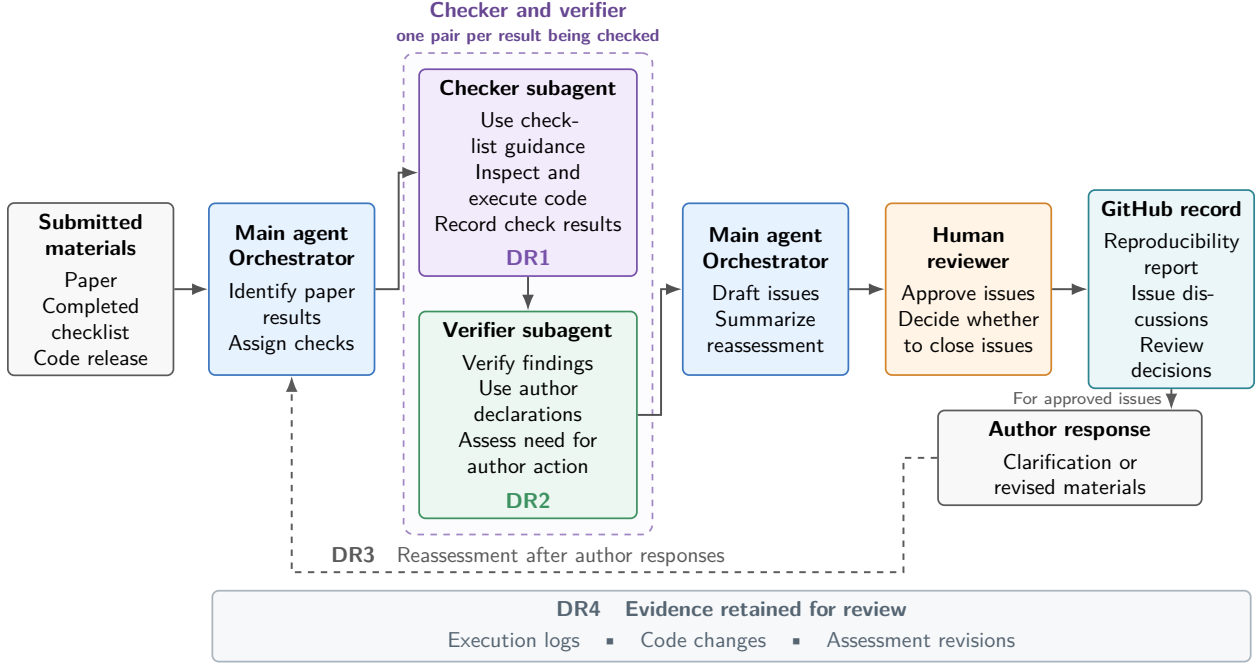
\begin{figure*}[t]
  \centering
  \begin{tikzpicture}[
    font=\sffamily,
    block/.style={
        rectangle,
        rounded corners=3pt,
        draw,
        line width=0.8pt,
        align=center,
        inner sep=5pt,
        font=\sffamily\small
    },
    outerBlock/.style={
        block,
        minimum width=2.70cm,
        text width=2.30cm,
        minimum height=2.80cm
    },
    workerBlock/.style={
        block,
        minimum width=3.55cm,
        text width=3.15cm,
        minimum height=3.25cm
    },
    inputBlock/.style={outerBlock, fill=inputFill, draw=inputDraw},
    orchBlock/.style={outerBlock, fill=orchFill, draw=orchDraw},
    checkerBlock/.style={
        workerBlock,
        fill=checkerFill,
        draw=checkerDraw
    },
    verifierBlock/.style={
        workerBlock,
        fill=verifierFill,
        draw=verifierDraw
    },
    humanBlock/.style={
        outerBlock,
        fill=humanFill,
        draw=humanDraw
    },
    outputBlock/.style={
        outerBlock,
        fill=outputFill,
        draw=outputDraw
    },
    panel/.style={
        rectangle,
        rounded corners=5pt,
        draw=checkerDraw!70,
        fill=checkerFill!22,
        dashed,
        line width=0.8pt
    },
    panelTitle/.style={
        font=\sffamily\bfseries\small,
        text=checkerDraw!85!black,
        align=center
    },
    arrow/.style={-{Latex[length=2.6mm]}, line width=0.9pt, draw=black!68},
    responseArrow/.style={arrow, dashed, draw=inputDraw}
]

\node[inputBlock] (inputs) at (0,0) {
    \textbf{Submitted materials}\\[3pt]
    Paper\\
    Completed checklist\\
    Code release
};

\node[orchBlock, right=0.55cm of inputs] (extract) {
    \textbf{Main agent}\\
    \textbf{Orchestrator}\\[3pt]
    Identify paper\\
    results\\
    Assign checks
};

\node[checkerBlock, right=0.70cm of extract, yshift=1.90cm] (checker) {
    \textbf{Checker subagent}\\[3pt]
    Use checklist guidance\\
    Inspect and\\
    execute code\\
    Record check results\\[4pt]
    \textcolor{checkerDraw}{\textbf{DR1}}
};

\node[verifierBlock, below=0.55cm of checker] (verifier) {
    \textbf{Verifier subagent}\\[3pt]
    Verify findings\\
    Use author\\
    declarations\\
    Assess need for\\
    author action\\[4pt]
    \textcolor{verifierDraw}{\textbf{DR2}}
};

\node[orchBlock, right=0.72cm of checker, yshift=-1.90cm] (merge) {
    \textbf{Main agent}\\
    \textbf{Orchestrator}\\[3pt]
    Draft issues\\
    Summarize\\
    reassessment
};

\node[humanBlock, right=0.58cm of merge] (human) {
    \textbf{Human reviewer}\\[3pt]
    Approve issues\\
    Decide whether\\
    to close issues
};

\node[outputBlock, right=0.58cm of human] (outputs) {
    \textbf{GitHub record}\\[3pt]
    Reproducibility report\\
    Issue discussions\\
    Review decisions
};

\begin{scope}[on background layer]
    \node[
        panel,
        fit=(checker)(verifier),
        inner xsep=0.24cm,
        inner ysep=0.25cm,
        label={[panelTitle]above:{Checker and verifier\\[-1pt]\scriptsize one pair per result being checked}}
    ] (workerPanel) {};
\end{scope}

\draw[arrow] (inputs) -- (extract);
\draw[arrow] (extract.east) -- ++(0.38,0) |- (checker.west);
\draw[arrow] (checker) -- (verifier);
\draw[arrow] (verifier.east) -- ++(0.38,0) |- (merge.west);
\draw[arrow] (merge) -- (human);
\draw[arrow] (human) -- (outputs);

\node[
    block,
    fill=inputFill,
    draw=inputDraw,
    minimum width=4.30cm,
    text width=3.90cm,
    minimum height=1.40cm
] (authors) at ($(human.center)!0.5!(outputs.center)+(0,-2.75cm)$) {
    \textbf{Author response}\\[3pt]
    Clarification or\\
    revised materials
};

\draw[responseArrow] (outputs.south)
    -- node[left, font=\sffamily\scriptsize, text=inputDraw, inner sep=4pt]
       {For approved issues}
    (outputs.south |- authors.north);

\coordinate (responseRoute) at ($(workerPanel.south)+(0,-0.60cm)$);
\draw[responseArrow] (authors.west) -- ++(-0.55cm,0)
    |- (extract.south |- responseRoute) -- (extract.south);
\node[
    font=\sffamily\small,
    text=inputDraw,
    fill=white,
    inner sep=2pt
] at ($(responseRoute)+(0,0.25cm)$) {
    \textbf{DR3}\quad Reassessment after author responses
};

\coordinate (recordCenter) at ($(extract.west)!0.5!(outputs.east)$);
\node[
    block,
    fill=recordFill,
    draw=recordDraw!55,
    minimum width=16.93cm,
    text width=16.45cm,
    minimum height=0.90cm,
    text=recordDraw!85!black
] at ($(recordCenter |- responseRoute)+(0,-0.88cm)$) {
    \textbf{DR4\quad Evidence retained for review}\\[2pt]
    Execution logs\quad\textbullet\quad Code changes\quad\textbullet\quad Assessment revisions
};

\end{tikzpicture}
  \caption{Agent architecture for reproducibility checking. DR labels mark
  where each requirement is primarily addressed. The dashed return path
  carries author responses and any revised materials into reassessment.
  Execution and assessment records are preserved across the process.
  Human reviewers approve issues and decide whether to close them.}
  \Description{The paper, completed checklist, and code release enter a main
  agent that identifies results and assigns checks. A checker uses checklist
  guidance to inspect and execute code and record the results. A verifier
  reviews findings and uses author declarations to assess the need for author
  action. The main agent drafts issues or summarizes reassessment. A human
  reviewer approves issues and decides whether to close them; reports,
  discussions, and decisions are retained on GitHub. For approved issues, a
  dashed path carries author clarifications or revised materials back to the
  main agent for reassessment by the checker and verifier. A band beneath the
  architecture shows that execution logs, code changes, and assessment
  revisions are retained for review across the process.}
  \label{fig:staix-agent-architecture}
\end{figure*}

Figure~\ref{fig:staix-agent-architecture} gives an overview of the system.
The main agent identifies experimental tables and figures in the paper's
main body and treats each as a \emph{result item}. Each result item is assigned
to a checker subagent for assessment as a single unit. The checkers share
a controlled GitHub Codespaces environment, with an isolated
workspace for each item. Given our computational resource constraints,
we limited each checker to 30 minutes per result item. A separate verifier subagent
decides whether further action from the authors is needed and corrects unsupported findings or vague result descriptions.
The main agent merges the verified assessments and prepares candidate GitHub issues awaiting human approval. 
For reassessment, the checker and verifier agents examine author responses and
updated materials before human reviewers decide whether to close an issue.
The following subsections explain how we address the four design requirements.

\subsection{DR1: Code Checks Should Use Author Declarations as Guidance}

\begin{figure}[t]
  \centering
  \includegraphics[width=\linewidth]{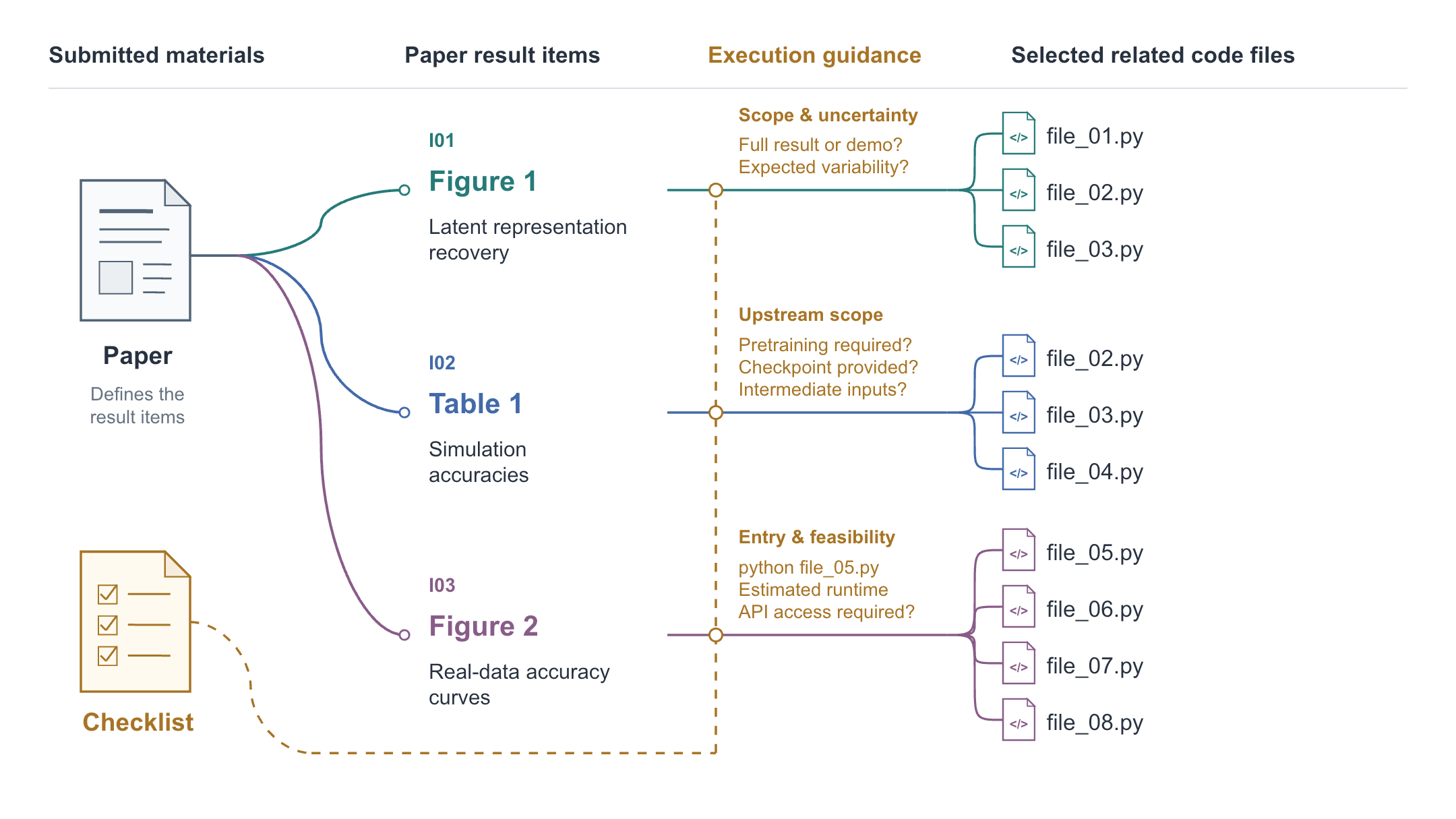}
  \caption{Using author declarations to guide checks of paper results.
  The guidance column illustrates checklist information about the scope and
  feasibility of the execution paths investigated.}
  \Description{Solid branches connect the paper to three result items and
  their associated Python files. A separate dashed route connects the
  checklist to illustrative questions about resource requirements, upstream
  computation, reproduction scope, and expected variability. These are example
  fields rather than completed answers from this submission. Repeated
  anonymous filenames indicate shared code.}
  \label{fig:dr1-code-mapping}
\end{figure}

Locating the code for a figure or table can be difficult in a large repository,
and running it may exceed the audit budget. We therefore design the agent to
use author declarations in the checklist to guide its search and plan feasible
checks. Figure~\ref{fig:dr1-code-mapping} shows how the agent links
each result item to relevant code files, using author declarations as a
starting point for investigation. After identifying the experimental tables
and figures in the main paper, it reads the corresponding checklist rows to
locate the declared code entry points and commands. The checker is also asked
to check whether the code is consistent with the paper's description of the experiment and the checklist's declarations, 
because sometimes the information in the checklist may be inaccurate. 

The checklist also helps the agent determine how much of each experiment the authors intend their submitted code to support. For example, the materials may support only a demonstration or an evaluation using a model checkpoint provided by the authors. We use these statements together with runtime estimates and resource requirements to guide the agent's execution plan within the audit budget. This allows author knowledge to inform both which code the agent runs and which parts of each experiment can be checked within the audit budget.

\Needspace{10\baselineskip}
\subsection{DR2: Agents Need to Consider Author Declarations When Interpreting Outcomes and Justifying Requests}
\label{sec:agent-interpretation}

\begin{figure}[tbp]
    \centering
    \includegraphics[width=\linewidth]{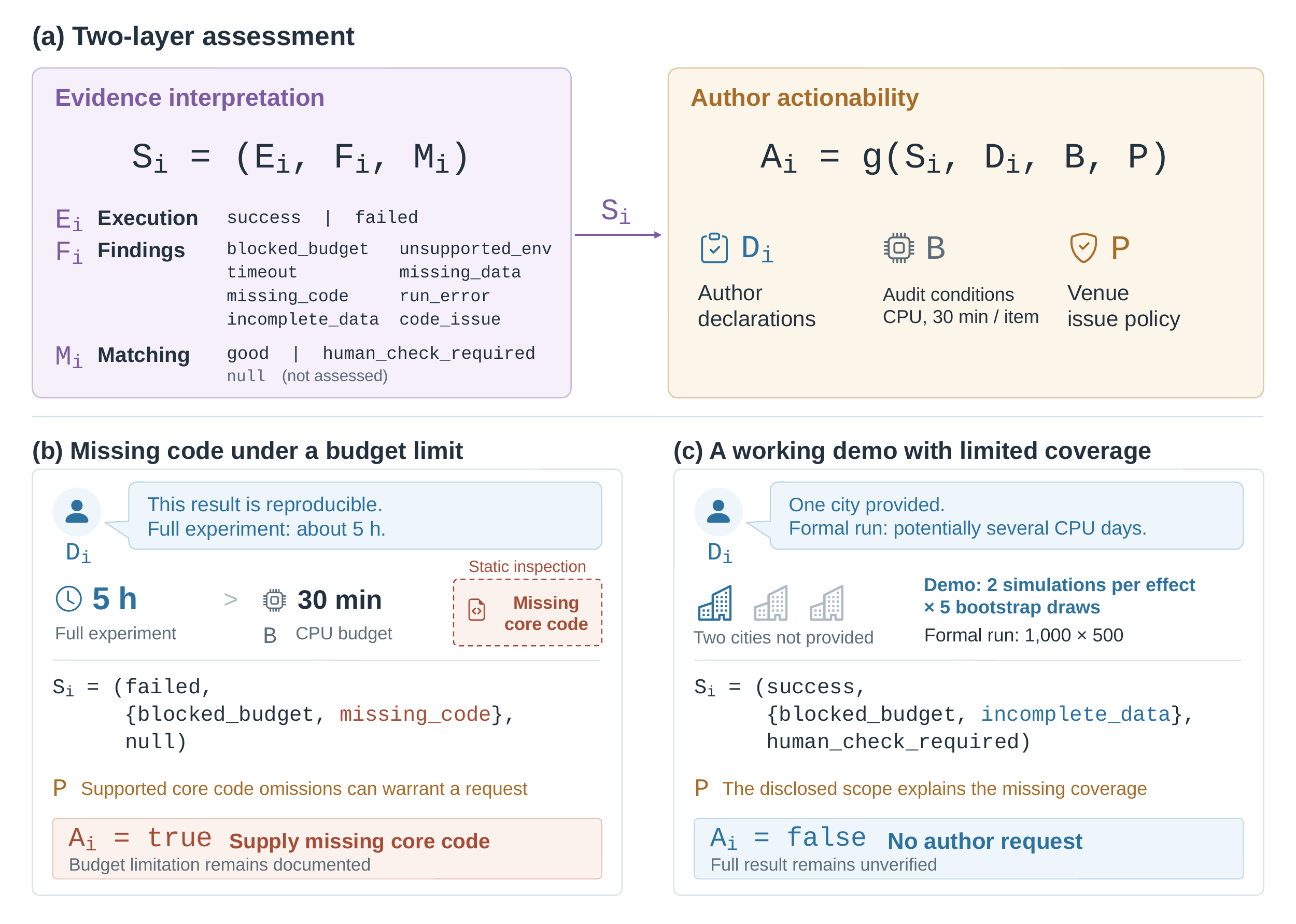}
    \caption{Assessing reproduction outcomes separately from the need for author action.
    (a) The agent evaluates audit evidence and separately decides whether author action is needed.
    (b) Full reproduction exceeds the audit's time budget, but the authors still need to supply missing core code.
    (c) The demo covers only one city, as the authors declared, so no further action is requested.}
    \Description{Panel a represents the evidence state S as execution E,
    findings F, and matching M, followed by an actionability judgment A informed
    by author declarations D, audit conditions B, and venue policy P.
    Execution is success or failed. Finding tags are blocked budget,
    unsupported environment, timeout, missing data, missing code, run error,
    incomplete data, and code issue. Matching is good, human check required,
    or null for an unassessed comparison. In panel b, the authors estimate
    five hours for a full experiment, exceeding the thirty-minute CPU audit
    limit. Static inspection separately establishes missing core code.
    The state is failed with blocked budget and missing code findings and
    null matching; actionability is true. Panel c shows a demo for one of
    three cities, using two simulations per effect and five bootstrap draws,
    compared with a formal setting of one thousand simulations per effect
    and five hundred bootstrap draws.
    The state is success with blocked budget and incomplete data findings
    and human check required for matching. The authors' disclosure explains
    the limited coverage, so actionability is false. The full reported result
    remains unverified.}
    \label{fig:dr3-evidence-actionability}
\end{figure}

When a check reveals a potential issue or an unexpected result, the agent needs to
determine what it means for reproducibility and whether the authors need to
address it. The interpretation depends on how the results were obtained and
what counts as agreement with the
paper~\cite{gundersen2021fundamental,baek2026automated,wu2026artifact}.
Resource limits can also leave parts of an experiment
untested~\cite{krafczyk2021learning,pallez2025implementing,delia2025lessons}.
We therefore design the agent to interpret reproduction outcomes in a three-dimensional state model and separately judge whether authors need to clarify or revise their materials.
Figure~\ref{fig:dr3-evidence-actionability}(a) illustrates these two layers.

The first layer represents the evidential state of result item $i$ as
\[
S_i=(E_i,F_i,M_i),
\]
with three fields:

\begin{itemize}
    \item \textbf{Execution ($E_i$): whether the computation produced output.}
    As shown in Figure~\ref{fig:dr3-evidence-actionability}(a),
    $E_i$ takes one of two values: \texttt{success} when the selected
    computation generates fresh numeric values or plot data, and
    \texttt{failed} otherwise. Success applies to the computation actually
    executed, including reduced experiments. Agreement with the paper is
    assessed separately through $M_i$.

    \item \textbf{Findings ($F_i$): what constrained or compromised the check.}
    A result item may have multiple findings, each recorded as a diagnostic
tag with a supporting reason. Figure~\ref{fig:dr3-evidence-actionability}(a)
lists all possible tags.

    \item \textbf{Matching ($M_i$): what comparison with the paper establishes.}
    Numeric results generated by agents are compared with the values reported in the paper,
    taking applicable tolerances stated by the authors into account. We do not
    rely solely on the agent to judge whether a reproduced figure matches the
    corresponding figure in the paper. Reproduced figures therefore default to
    \texttt{human\_check\_required}. For every result with this label, a human
    reviewer checks whether it matches the corresponding result in the paper.
    The value \texttt{null} indicates that execution did
    not succeed, so agreement with the paper was not assessed.
\end{itemize}

The second layer determines whether the evidential state $S_i$ warrants a request to
the authors:
\[
A_i=g(S_i,D_i,B,P),
\]
where $D_i$ denotes the authors' result item-specific declarations, $B$ the audit
budget and execution conditions, and $P$ the venue's reproducibility policy.
The agent first checks whether the authors' declarations explain the specific
limitation or discrepancy observed. It raises an issue only
if the evidence still shows a significant problem that the authors can address.
For example, an author's statement
that only one dataset is provided may explain limited coverage so no further action is required. Conversely, if authors state that
the required code is included but the implementation is absent, this would support the agent to create an open GitHub issue.
The complete assessment, $R_i=(S_i,A_i)$, distinguishes between interpreting
reproduction outcomes and deciding whether authors need to
revise their materials.

\Needspace{6\baselineskip}
The two initial audit cases in Figure~\ref{fig:dr3-evidence-actionability}
illustrate why this distinction matters: full reproduction remained
unverified in both, but only one warranted a request to the authors.
In (b), the authors' five hour runtime
estimate exceeded the checking budget, but the checker found the core code is missing. This code omission warranted a request
even though full execution remained out of reach. In (c), a working demo
established only part of the paper's results; the authors had explicitly disclosed the
restricted data coverage. That disclosure explained the missing coverage
without establishing the full result, allowing the limitation to remain
recorded without an open issue.

During the auditing procedure, the checker agent first proposes $S_i$. The verifier agent can revise $S_i$, checking whether the checker's findings are supported and whether the corresponding reasons are clear. 
The verifier agent then determines $A_i$ based on $S_i$, the authors' declarations, and the venue's policy. Human reviewers then use the verified assessments to decide which requests to approve before they are sent to authors.

\Needspace{8\baselineskip}
\subsection{DR3: Reassessment Requires Checking Author Responses}

\begin{figure}[!htbp]
    \centering
    \includegraphics[width=\linewidth]{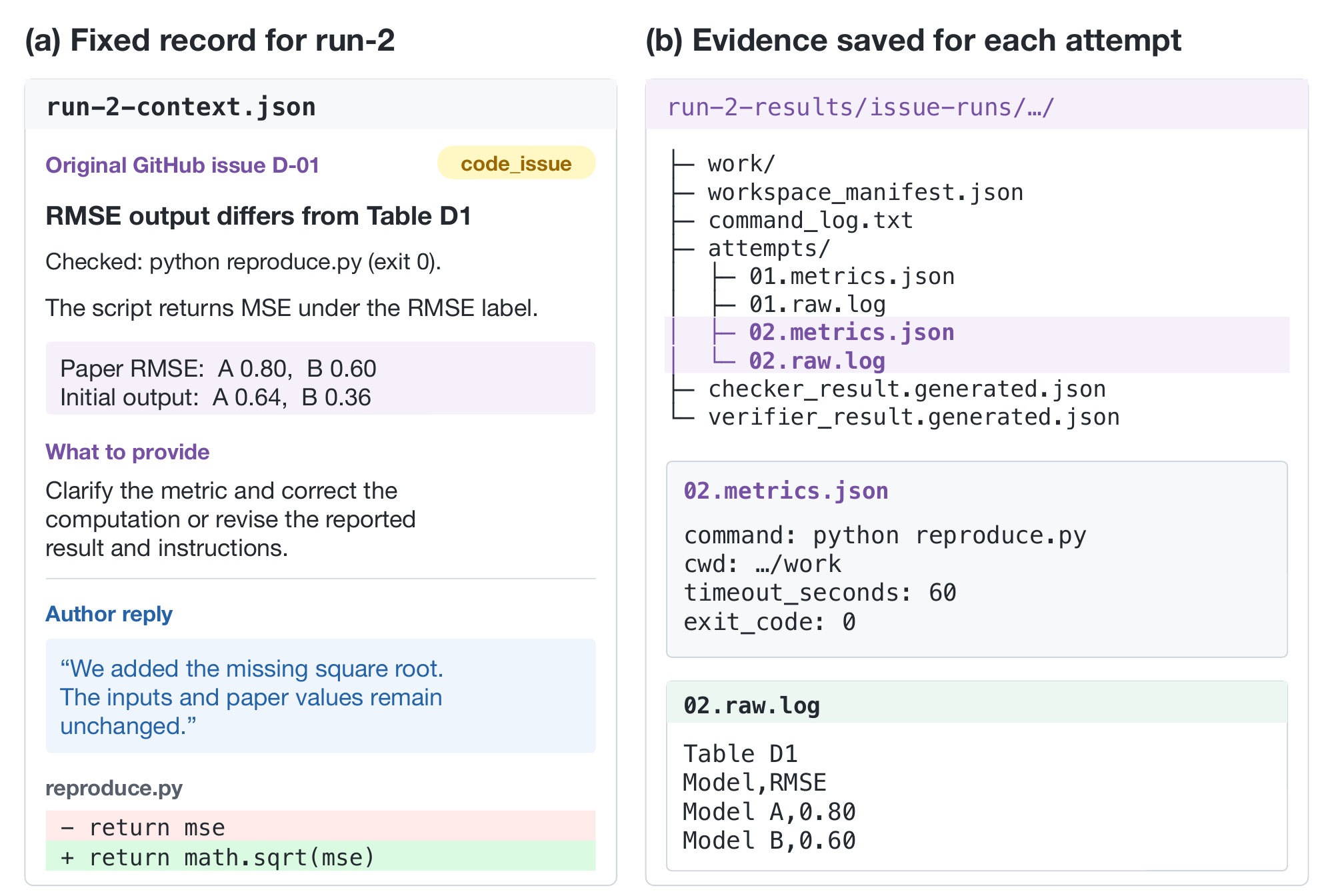}
    \caption{Records supporting run-2 and auditability, continuing the
    synthetic Table~D1 example in Figure~\ref{fig:case-a-collaborative-audit}.
    (a) A fixed record links the original GitHub issue and findings to the
    author reply and revised code. (b) Each issue has a workspace with numbered
    execution records and raw output, alongside separate checker and verifier
    assessments. The expanded attempt shows a local execution of the corrected
    synthetic code. }
    \Description{Two panels show run-2 inputs and retained execution
    evidence. On the left, a human-readable view of run-2-context.json links
    issue D-01 to an author reply and code revision. The issue includes its
    title, a code issue tag, the executed command, the output mismatch, and a
    request to clarify or correct the metric. The original script produced
    MSE values of 0.64 and 0.36 under an RMSE label. The reply reports adding
    a square root, shown in the code change. On the right, a directory tree shows working
    files, an input manifest, command history, numbered metrics and raw logs,
    and separate checker and verifier records. The expanded attempt 02 shows
    the command, working directory, timeout, and exit code. Its output contains
    RMSE values of 0.80 and 0.60. The code
    execution is real; the author interaction and assessment records are
    illustrative.}
    \label{fig:dr3-dr4-audit-record}
\end{figure}

An author response may report a repair or reveal that the agent misunderstood
the submitted code. We therefore design the run-2 agent to check the authors' responses and any revisions to determine whether further author action is needed. Before reassessment, we
prepare a fixed record linking each issue and the agent's findings during run-1 to the
author's replies and the versions of any revised materials.
Figure~\ref{fig:dr3-dr4-audit-record}(a) illustrates this record using the
synthetic example from Figure~\ref{fig:case-a-collaborative-audit}.
Working from this record, the checker agent compares relevant files with the original
submission and tests the authors' explanations against the code and data. If
these checks show that an earlier finding was mistaken, the checker no longer
treats it as an unresolved problem. The verifier agent independently reviews the
evidence and recommends closing
or keeping the issue open, pointing out cases where the evidence is insufficient to
decide. Human reviewers make the final decision on closure and approve any
further requests to authors.

\subsection{DR4: Agent Checks and Judgments Must Remain Auditable}

To evaluate an agent's recommendation, authors and reviewers need to see how
it checked the code and obtained its evidence. We therefore retain the
command and raw output for each numbered execution attempt, as
illustrated in Figure~\ref{fig:dr3-dr4-audit-record}(b). The agent works on a
copy of the submitted code, leaving the original unchanged. Any code changes
are recorded with the corresponding execution attempt number. Later attempts
leave earlier records intact.

The verifier's assessment is saved separately from the checker's original
judgment. When the verifier changes the checker's assessment, it records the
correction and its reason, allowing reviewers to compare the two assessments.

\section{Findings}
\label{sec:findings}
\label{sec:findings-coverage}

In this section, we draw on the conference deployment and comparative experiments to address the three research questions described in Section~\ref{sec:introduction}.

The official run-1 audit covered 74 figures and 59 tables, each reviewed
by a checker agent and a verifier agent. For figure results, the agent left matching
judgments to human reviewers, who compared every generated plot with the
corresponding figure in the paper. All plots from successful executions
without a \texttt{blocked\_budget} tag were judged to be good matches.
The agents spent a total of 6.34 hours running commands during the run-1 audit.
The ten papers with open GitHub issues underwent run-2 reassessment.
Remaining concerns were then addressed through further author clarification
or revision and discussion with reviewers.

To investigate whether the checklist provided useful information for the
venue's reproducibility checks, we compared audits using complete
checklists with audits using blank templates. We examined differences in
how the agent investigated the code and judged the need for author action.
The experiment covered 39 result items from the same ten papers, with three
repeats per condition. Table~\ref{tab:audit-data-sources} summarizes the
audit configurations and main findings. The following subsections, we will present
these findings in detail to address the three research questions introduced
in Section~\ref{sec:introduction}.

\begin{table}[htbp]
  \caption{Audit configurations and main findings.}
  \label{tab:audit-data-sources}
  \centering
  \small
  \setlength{\tabcolsep}{3.5pt}
  \renewcommand{\arraystretch}{1.15}
  \begin{tabular}{@{}p{0.16\linewidth}p{0.18\linewidth}p{0.15\linewidth}p{\dimexpr0.37\linewidth-8\tabcolsep\relax}p{0.14\linewidth}@{}}
    \toprule
    \raggedright\textbf{Audit records} & \raggedright\textbf{Coverage} & \raggedright\textbf{Recorded model} & \raggedright\textbf{Findings} & \raggedright\textbf{Average token usage} \tabularnewline
    \midrule
    \raggedright Official initial audit (run-1) &
    \raggedright 36 papers; 133 result items &
    \raggedright GPT-5.5 &
    \raggedright The run-1 audit yielded 20 open GitHub issues after human approval. &
    \raggedright Not available \tabularnewline
    \addlinespace
    \raggedright Official reassessment (run-2) &
    \raggedright 10 papers with 20 original GitHub issues &
    \raggedright GPT-5.6-Sol &
    \raggedright The run-2 audit recommended closing 15 issues. All 20 were ultimately closed following author responses, further checks, and human review. &
    \raggedright 13.09 million \tabularnewline
    \addlinespace
    \raggedright Checklist comparison experiment &
    \raggedright The same 10 papers; three repeats with or without checklists &
    \raggedright GPT-5.6-Sol &
    \raggedright Audit records showed how declarations informed investigation planning and judgments about author action. Complete checklists yielded fewer requests overall. &
    \raggedright 25.13 million \tabularnewline
    \bottomrule
  \end{tabular}
  \par\smallskip
  \parbox{\linewidth}{\footnotesize
    All configurations specify \texttt{xhigh} reasoning.
    Token usage is the mean total input and output tokens per paper audit
    across the orchestrator and all checker and verifier sessions. Original run-1 token logs were unavailable.}
\end{table}

\subsection{RQ1. How can a venue structure collaboration between authors
and reviewers to reduce information asymmetry in reproducibility auditing?}
\label{sec:findings-rq1}
\label{sec:findings-declarations}

To reach agreement on reproducibility, authors and the venue need a clear shared standard and an efficient way to collaborate. Our checklist makes expectations
explicit through author declarations, while the two-stage collaborative workflow
coordinates authors, agents, and human reviewers under a transparent environment. In this subsection, we first examine whether the checklist provides additional information from authors that informs the agent's judgments about the need for author action, then discuss how the collaborative workflow supports efficient and transparent communication between authors and reviewers.

\paragraph{The role of structured author declarations.}
\label{sec:findings-checklist-comparison}
Author self-attestations supplied a concrete reference for assessing what
each submitted release was expected to support. In the official run-1,
125 of 133 final agent outputs (94.0\%) explicitly cited the checklist in the audit record, and all 20 published GitHub issues referred to an author
statement in the checklist. These records show that author declarations entered the assessment
process, while it remains unknown how agent judgments would differ without the checklist.

We therefore examined this question through an experiment on those ten papers that
had received approved issues. For each
paper, we ran the run-1 procedure using a complete checklist compared
with using a blank copy of the template, with the remaining research
materials unchanged. We repeated the experiment three times to reduce the effect
of stochasticity in the agent's reasoning. Both conditions used
the same run-1 assessment prompts and
actionability policy.

As shown in Table~\ref{tab:checklist-actionability}, the agent judged more
result items to require author action with blank checklists in every repeat.
Across three repeats of the same 39 result items, the agent judged author
action necessary in 68 of 117 assessments (58.1\%) with complete checklists,
compared with 84 of 117 (71.8\%) with blank checklists, a difference of
13.7 percentage points.\footnote{The official run-1 audit yielded 20 published
GitHub issues. In the three complete checklist reruns, the agent judged
22, 25, and 21 result items to require author action, respectively.}
The different judgments with and without checklists suggest that the agent
used information supplied by authors when deciding whether further action was needed. This provides evidence that the checklist helped
bridge the information gap between authors and the venue by bringing
authors' knowledge into the audit process.
Section~\ref{sec:findings-rq2} further uses detailed analysis to examine
how the checklist shaped the agent's decisions.

\begin{table}[tbp]
  \caption{Number of result items judged by the agent to require author action
  in each condition. Each repeat checks the same 39 result items from ten papers
  with a complete or blank checklist. Difference is the count with a blank
  checklist minus the count with a complete checklist.}
  \label{tab:checklist-actionability}
  \centering
  \small
  \begin{tabular}{@{}lrrr@{}}
    \toprule
    Repeat & Complete checklist & Blank checklist & Difference \\
    \midrule
    1 & 22 & 28 & $+6$ \\
    2 & 25 & 29 & $+4$ \\
    3 & 21 & 27 & $+6$ \\
    \bottomrule
  \end{tabular}
\end{table}

\paragraph{Efficiency and transparency of the conference's collaborative workflow.}
The agent completed each of the 30 repeat audits with complete checklists
in under an hour, with a median of 40.1 minutes per paper (range 17.9--57.6).
The ten final official run-2 reassessments also finished within an hour per
paper, with a median of 21.9 minutes (range 18.1--51.2).

One discussion on GitHub showed how an author and a reviewer could quickly resolve a remaining concern through a focused request for clarification. After run-2, a reviewer asked an author
to identify the configuration and label files corresponding to a supplied
demonstration, explicitly stating that a full rerun was unnecessary. The
author replied within six hours, enabling the reviewer to confirm the files
and run the demonstration with minor adjustments. The issue was closed within
15 hours of the request. In practice, not all cases were resolved this quickly as we gave authors ample time to respond. Even so, this example suggests that the workflow can make collaboration between authors and reviewers straightforward, allowing them to resolve a remaining concern through a brief, focused discussion.

GitHub platform also makes the collaboration process transparent by allowing authors and reviewers to see what evidence supported each decision. In each paper, the audit report and GitHub issue threads showed explicitly what authors were asked to clarify or fix with corresponding evidence and comments. Within each
paper's repository, all the agent checking history and decisions are visible to both authors and reviewers. Each GitHub issue thread explained why the agent requested author action
and provided supporting commands and comparisons with the paper's results.
Authors responded in the same thread, where reviewers assessed their replies
and explained the decision to close the issue or keep it open. This shared
record allowed both sides to trace the reasoning from the initial finding
to the final decision.

\subsection{RQ2. How can an AI agent use author declarations and responses
to interpret complex reproduction evidence and make auditable judgments about
the need for clarification or repair?}
\label{sec:findings-rq2}
\label{sec:findings-actionability}
\label{sec:findings-verification}
\label{sec:findings-consistency}

To address RQ2, we designed the agent described in Section~\ref{sec:agent-design}
to use author declarations and responses to guide investigation and the
interpretation of reproduction evidence. The checker agent separately records whether code execution succeeded, what problems were identified, and how the outputs compared with the paper's reported results. A verifier agent then uses these records to judge whether authors need to provide clarification or make repairs under the venue's audit conditions. Here we investigate how the agent make use of author's information. The traceable audit records allow us to examine how agents
used author declarations in practice. By retrospectively examining the official run-1 audit and the checklist comparison experiment record in
Section~\ref{sec:findings-rq1}, we found that author declarations informed
agents' decisions about what to check and how to interpret the resulting
evidence.

\paragraph{Author declarations guided investigation planning.}
After manual review, we found that for 37 of the 39 result items in the official run-1 assessments of these ten
papers that have open GitHub issues, the assessment by our agent explicitly linked author declarations to investigation
planning. The assessments cited author declarations to explain which code
and inputs the agents chose, what they checked, and why they reduced or
skipped some computations.

In the comparison experiment, we found that runtime estimates stated in the checklist helped agents plan checks within
the audit budget. For example, one
checklist estimated six to ten hours for upstream inference and supplied a
separate downstream analysis command. All three agent runs with a complete
checklist cited this estimate and skipped full inference because it would
exceed the 30-minute audit budget. They still attempted the specified
downstream analysis, but found that the required script was missing. These agents treated the time limit as a valid reason to skip full inference, but still asked the authors to provide the missing analysis script.

However, following the authors' specified reproduction path could also narrow the investigation. In another paper, the checklist specified that a table should be assembled from saved experimental results. All three agent runs with a complete checklist followed this path and explicitly cited it as the reason for not rerunning the experiments. Without a checklist specifying where to start, one agent traced the table values back to the supplied datasets and inspected the code used to generate them. Although recalculating the values from these datasets reproduced the table, the agent discovered that the data generation code used a different threshold from the one described in the paper. It therefore requested author action.

These cases show how author checklists helped agents choose what code to run within the audit budget. However, this guidance sometimes could also narrow the investigation to the declared reproduction path, leaving problems elsewhere in the code undetected.

\paragraph{Author declarations informed the interpretation of findings.}
As described in Section~\ref{sec:agent-interpretation}, the verifier agent reviews code
execution outcomes, diagnostic findings, and comparisons with the paper,
then considers author declarations and the venue's audit conditions to
judge whether clarification or repair is warranted. Our manual review of the
audit records confirmed that the agent explicitly used author declarations
to justify requesting or withholding author action for 21 of the same 39
result items.

Author declarations give authors an opportunity to explain reproduction limitations and make the case that no further action should be required of them. Therefore, we would expect fewer judgments requiring
author action when agents have complete checklists. The comparison in
Section~\ref{sec:findings-rq1} was consistent with this expectation: across
the three audits per condition, 13 of the 39 result items received fewer such
judgments with complete checklists than with blank checklists, and 23 received
the same number. However, three items received more judgments requiring author
action with complete checklists, which is counterintuitive -- author commitments could themselves provide grounds for an additional repair
request. For one figure, the checklist
specified a script for calculating the plotted metrics from saved model
predictions. All three audits with a complete checklist found that this script
was missing and asked the authors to provide it as promised. A blank-checklist
audit identified the same omission but emphasized that the model's training
and prediction code was already present. It regarded the missing metric
calculations and plotting as additional analysis beyond that core
implementation and did not request repair.

Through these cases, we see that our agent design explicitly make use of checklist information, showing that the need of author's information when judging whether a potential issue is actionable. We also need to be careful that sometimes checklist can ``steer'' agent's behavior and make the agent oversee some potential issues. In the next section, we will discuss our finding statistics during the conference deployment.

\subsection{RQ3. What reproducibility problems does the pipeline identify
in a conference deployment, and to what extent are they resolved after author
responses and further checks?}
\label{sec:findings-rq3}
\label{sec:author-responses}
\label{sec:run2-results}

The conference audit records allow us to trace how individual problems were handled after they were raised with authors. We connect the initial findings to authors' revisions and clarifications, then examine how subsequent agent checks and human review determined whether each concern had been addressed.

\begin{figure}[tbp]
  \centering
  \includegraphics[width=\textwidth]{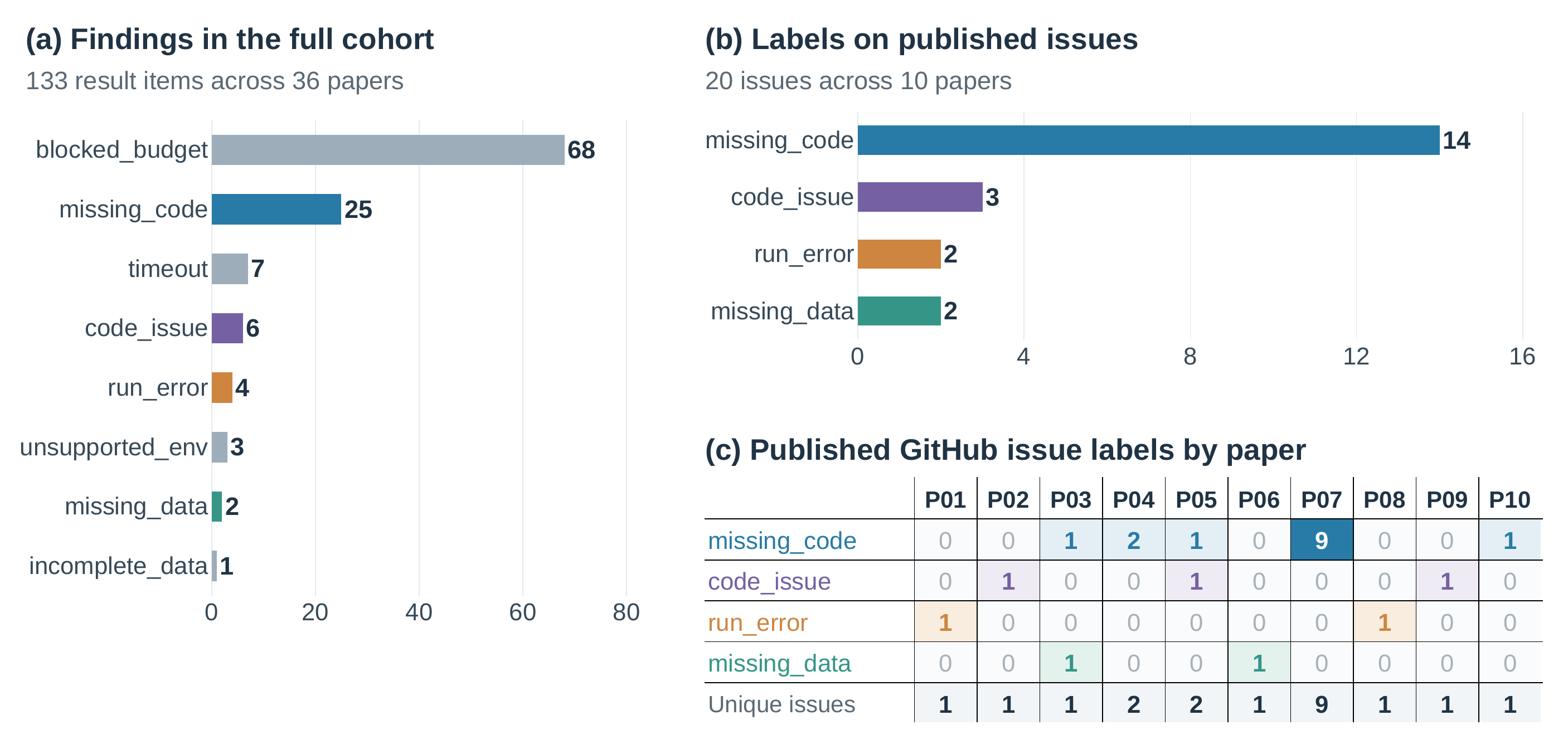}
  \caption{Findings and published issues in the initial audit.
  (a) Finding tags across 133 result items from 36 papers, including
  findings that did not warrant author action.
  (b) Labels on the 20 published issues.
  (c) Their distribution across ten anonymized papers, with unique issue
  counts shown separately. Each issue concerns one result item.
  Tags and labels can overlap: one published
  issue carries both \texttt{missing\_code} and \texttt{missing\_data},
  giving 21 issue labels in total.}
  \Description{Two bar charts and a matrix summarize the initial audit.
  Budget limitations occur in 68 result items and missing code in 25.
  Published issues carry 14 missing-code labels, three code-issue labels,
  two run-error labels, and two missing-data labels. Paper P07 accounts for
  nine missing-code issues. Paper P03 has one issue carrying two labels.}
  \label{fig:open-issue-label-distribution}
\end{figure}

\paragraph{What reproducibility problems were identified.}
Figure~\ref{fig:open-issue-label-distribution} summarizes the findings across
133 result items and the issues raised with authors. The audit identified
\texttt{missing\_code}, \texttt{missing\_data}, \texttt{run\_error}, and
\texttt{code\_issue} findings. The most frequent tag was
\texttt{blocked\_budget} (full reproduction was infeasible within the available
time or resources), recorded for 68 items across 27 papers.
The agent's audit reports identified problems in the submitted code or data. They also recorded cases where the audit could not complete reproduction because of limited time or resources.

We quantified how execution outcomes related to the agent's judgments
about author action. Of the 85 items with execution status
\texttt{failed}, 68 were judged not to warrant author action.
Conversely, three of the 48 items with execution status \texttt{success}
still warranted action due to mismatched or incomplete outputs.
Successful execution therefore did not guarantee agreement with the paper,
consistent with prior work that distinguishes running code from reproducing
results~\cite{baek2026automated,wu2026artifact}.
Budget constraints also did not rule out actionable problems: six of the
68 items tagged \texttt{blocked\_budget} contained code or data problems
judged to warrant clarification or repair. These patterns show the importance of separately assessing whether a result has been reproduced and whether the evidence warrants author action. Incomplete verification alone neither confirms an error nor
rules out a problem requiring author action.

Following human review, 20 GitHub issues were published across ten papers.
The \texttt{missing\_code} label appeared in 14 issues from five papers,
with nine concentrated in a single submission. Other labels
included \texttt{code\_issue}, \texttt{run\_error}, and
\texttt{missing\_data}.

\paragraph{What responses and changes authors reported.}
All ten author teams that received issues reported updating their code.
Table~\ref{tab:author-response-actions} summarizes the changes and
clarifications described in their responses. For example, the authors of P09 added the missing code to compute coverage and mean interval width. They also explained how
each method constructed the intervals used to calculate these metrics.

Authors of five papers clarified which results could be reproduced from
the submitted materials or how to interpret the outputs. In P07, the
authors acknowledged that several scripts listed as available in the
checklist were missing from the submitted code. They replied in GitHub comment that
 they have supplied these scripts and corrected the checklist to clarify which materials were available.

\begin{table}[H]
  \caption{Changes and clarifications reported by authors of the ten papers
  with open GitHub issues.
  Each paper is counted once per category, and categories may overlap.}
  \label{tab:author-response-actions}
  \centering
  \begin{tabular}{@{}p{0.78\linewidth}r@{}}
    \toprule
    Reported action & Papers \\
    \midrule
    Updated or completed code & 10 \\
    Added input datasets or example data & 4 \\
    Supplied precomputed experimental outputs & 1 \\
    Revised the manuscript & 2 \\
    Corrected the reproducibility checklist & 1 \\
    Clarified reproduction scope or interpretation of outputs & 5 \\
    \bottomrule
  \end{tabular}
\end{table}

\paragraph{Issue resolution after author responses.}
After reviewing authors' responses and updated materials, the run-2 agent
recommended closing 15 of the 20 original issues and keeping five open.
In nine of the 15 cases, the agent successfully ran the code. In the other
six, the agent recommended to close the issue after confirming that the missing code
had been supplied, with full reproduction still unverified because of
budget limits or unavailable inputs.
Table~\ref{tab:issue-resolution-paths} summarizes how all 20 original
concerns ultimately reached closure.

\begin{table}[tbp]
  \caption{Closure pathways for the 20 original concerns across ten papers.}
  \label{tab:issue-resolution-paths}
  \centering
  \renewcommand{\arraystretch}{1.15}
  \begin{tabular}{@{}p{0.78\linewidth}r@{}}
    \toprule
    Path to closure & Concerns \\
    \midrule
    Agent recommended closure after successful execution & 9 \\
    Agent recommended closure after confirming that missing code had been supplied & 6 \\
    Human review supported closure without further author replies & 2 \\
    Further author clarification and reviewer checks supported closure & 3 \\
    \midrule
    \textbf{Total} & \textbf{20} \\
    \bottomrule
  \end{tabular}
\end{table}

Human review supported closing two of the five issues that the agent
recommended keeping open, without additional author replies after run-2.
In P05, the reviewer confirmed that the runtime
figure could be redrawn from the authors' supplied measurements and
accepted that not all timings had been independently measured again.
In P07, the reviewer withdrew a run-2 \texttt{missing\_code} finding after
confirming that the calculation was already implemented elsewhere in the
submitted code. These cases show how human review corrected the agent's
assessment and determined which reproduction limits were acceptable under
the venue's audit conditions.

The remaining three issues were closed after further author clarification
and reviewer checks. Ultimately, all 20 original concerns were considered to be resolved, and the corresponding GitHub issues were closed. 

\section{Discussion and Conclusion}
\label{sec:discussion}
\label{sec:conclusion}

Our system design and conference deployment offers an initial view of how author participation can
support agentic reproducibility checking during peer review. However, the study's
scale and checking conditions limit how far these findings can be
generalized.

\paragraph{Limitations.}
Our study has several limitations. Resource constraints limited the depth
of checking for 68 of the 133 result items, leaving full reproduction
unverified even when reduced executions succeeded. Evidence about author
responses also comes from only ten author teams who received approved
issues at a single conference. Audit stability remains another concern.
Across three independent repetitions of the initial audit on the same ten
papers with complete checklists, the agent judged 22, 25, and 21 result
items to require author action. These differences indicate that its
assessments were not fully consistent. Audit outcomes may vary because the
agent examines different code, runs different computations, or interprets
the evidence differently. Our data do not show how much each of these
factors contributes to the variation. Checklist
guidance can also be interpreted too narrowly: in one reassessment, the
agent treated a calculation absent from the declared analysis script as
missing code despite its implementation elsewhere in the repository.
Future work could aim to improve audit consistency and make checklists easier to complete accurately. We will also explore ways to improve collaboration among authors, agents, and reviewers.

Despite these limitations, our work offers practical insights into how agentic reproducibility auditing can support peer review.

\paragraph{Implications for peer review.}
First, we want to emphasize that keeping authors-in-the-loop matters a lot in agentic
reproducibility auditing. Building on
collaboration between authors and reviewers~\cite{nuest2021codecheck,jasa2024partnering},
we designed a structured checklist and a two-stage collaborative workflow
on GitHub to create a transparent and efficient environment for
reproducibility checking and reduce information asymmetry between authors
and the venue. Our reproducibility checking agent uses checklist
declarations and subsequent author responses to interpret varied and
complex code execution outcomes and provide actionable feedback.
Comparative experiments and retrospective analysis of the official audits
further show that author declarations can influence the agent's
investigation planning and interpretation of code execution outcomes.
We in this paper provide a collaborative workflow and agent design for more
efficient reproducibility auditing, and show that information from authors
matters during both code checking and subsequent collaboration.

Second, our system has been deployed in real-world conference setting.
We quantified how code execution outcomes related to judgments about
author action. The findings highlight the importance of separately
assessing whether a result has been reproduced and whether the evidence
warrants clarification or repair. During the deployment, our agents identified missing code,
execution errors, and inconsistencies between the submitted code and the
paper. Authors responded with revised code, additional materials, and
clarification of reproduction conditions. All the issues were marked resolved after further agent checking and communication with our reviewer.
It is also worth pointing out that human oversight can be crucial, as agents can make mistakes and their values aren't well aligned with domain standards. Overall, our deployment shows that our author-in-the-loop agentic reproducibility auditing can help identify and resolve reproducibility problems in a conference setting.
Future deployments could develop more author- and agent- friendly checklist structure, or explore new ways to strengthen the communications between author and venue. One could also examine how to improve the consistency of agent assessments across repeated audits and scale the procedure to larger conferences. Such deployments could track both assessment consistency and the time required from authors and reviewers, helping establish whether the process can handle more submissions without increasing judgment variability or placing excessive demands on participants.

\bibliographystyle{plainnat}
\bibliography{references}

\end{document}